\documentclass{article}
\pdfoutput=1 

\usepackage{aaai22}  \usepackage[pdftex]{graphicx}
\usepackage{times}  \usepackage{helvet}  \usepackage{courier}  \usepackage[hyphens]{url}  \usepackage{graphicx}    \usepackage{natbib}  \usepackage{caption} \DeclareCaptionStyle{ruled}{labelfont=normalfont,labelsep=colon,strut=off}       \usepackage{algorithm}
\usepackage{algorithmic}

\usepackage{newfloat}
\usepackage{listings}
\floatstyle{ruled}
\newfloat{listing}{tb}{lst}{}
\floatname{listing}{Listing}
\usepackage[utf8]{inputenc} \usepackage[T1]{fontenc}    \usepackage{url}            \usepackage{booktabs}       \usepackage{amsfonts}       \usepackage{nicefrac}       \usepackage{microtype}      \usepackage{xcolor}         

\usepackage{booktabs} \usepackage{soul,color}
\usepackage{nicefrac}
\usepackage{subcaption}
\usepackage{multirow}
\usepackage{sidecap}
\usepackage{amsmath, latexsym}
\usepackage{algorithm}
\usepackage{algorithmic}
\usepackage{pifont}
\newcommand{\cmark}{\ding{51}}\newcommand{\xmark}{\text{\ding{55}}}
\usepackage{enumitem}
\usepackage{footnote}
\makesavenoteenv{tabular}
\makesavenoteenv{table}

\usepackage{xspace}
\usepackage[usestackEOL]{stackengine}

\usepackage[british,american]{babel}

\usepackage[pagebackref=true,breaklinks=true,letterpaper=true,bookmarks=false]{hyperref}

\usepackage[bottom]{footmisc}

\newcommand{\topic}[1]{\noindent \textbf{#1}}

\newcommand{\OURS}{NesT\xspace}

\usepackage{bibentry}

\begin{document}
\title{Nested Hierarchical Transformer: Towards Accurate, Data-Efficient and Interpretable Visual Understanding}
\author{
Zizhao Zhang$^{1}$ 
\;
Han Zhang$^{2}$  
\; 
Long Zhao$^{2}$ 
\;
Ting Chen$^{2}$ 
\;
Sercan \"{O}. Ar{\i}k$^{1}$
\; 
Tomas Pfister$^{1}$
}
\affiliations{
 $^{1}$Google Cloud AI \quad $^{2}$Google Research
}

\maketitle

\begin{abstract}
Hierarchical structures are popular in recent vision transformers, however, they require sophisticated designs and massive datasets to work well. 
In this paper, we explore the idea of nesting basic local transformers on non-overlapping image blocks and aggregating them in a hierarchical way. 
We find that the block aggregation function plays a critical role in enabling cross-block non-local information communication.
This observation leads us to design a simplified architecture that requires minor code changes upon the original vision transformer.
The benefits of the proposed judiciously-selected design are threefold: 
(1)~\OURS converges faster and requires much less training data to achieve good generalization on both ImageNet and small datasets like CIFAR;
(2)~when extending our key ideas to image generation, \OURS leads to a strong decoder that is 8$\times$ faster than previous transformer-based generators; and
(3)~we show that decoupling the feature learning and abstraction processes via this nested hierarchy in our design enables constructing a novel method (named GradCAT) for visually interpreting the learned model.
Source code is available \url{https://github.com/google-research/nested-transformer}.
\end{abstract}

\section{Introduction}

Vision Transformer (ViT) \cite{dosovitskiy2020image} model and its variants have received significant interests recently due to their superior performance on many core visual applications \cite{cordonnier2019relationship,liu2021swin}. 
ViT first splits an input image into patches, and then patches are treated in the same way as tokens in NLP applications.
Following, several self-attention layers are used to conduct global information communication to extract features for classification. 
Recent work~\cite{dosovitskiy2020image,cordonnier2019relationship} shows that ViT models can achieve better accuracy than state-of-the-art convnets~\cite{tan2019efficientnet,he2016deep} when trained on datasets with tens or hundreds of millions of labeled samples. 
However, when trained on smaller datasets, ViT usually underperforms its counterparts based on convolutional layers.
Addressing this data inefficiency is important to make ViT applicable to other application scenarios, e.g. semi-supervised learning \cite{sohn2020fixmatch} and generative modeling \cite{goodfellow2014generative,zhang2019self}. 

Lack of inductive bias such as locality and translation equivariance, is one explanation for the data inefficiency of ViT models.  
~\citet{cordonnier2019relationship} discovered that transformer models learn locality behaviors in a deformable convolution manner~\cite{dai2017deformable}: bottom layers attend locally to the surrounding pixels and top layers favor long-range dependency. 
On the other hand, global self-attention between pixel pairs in high-resolution images is computationally expensive. 
Reducing the self-attention range is one way to make the model training more computationally efficient \cite{beltagy2020longformer}.
These type of insights align with the recent structures with local self-attention and hierarchical transformer \cite{han2021transformer,vaswani2021scaling,liu2021swin}. 
Instead of holistic global self-attention, these perform attention on local image patches. 
To promote information communication across patches, they propose specialized designs such as the ``haloing operation''~\cite{vaswani2021scaling} and ``shifted window''~\cite{liu2021swin}. These are based on modifying the self-attention mechanism and often yields in complex architectures. Our design goal on the other hand keeping the attention as is, and introducing the design of the aggregation function, to improve the accuracy and data efficiency, while bringing interpretability benefits.

The proposed \OURS model stacks canonical transformer blocks to process non-overlapping image blocks individually. 
Cross-block self-attention is achieved by nesting these transformers hierarchically and connecting them with a proposed aggregation function.
Fig. \ref{fig:arch} illustrates the overall architecture and the simple pseudo code to generate it. Our contributions can be summarized as:
\begin{figure*}[t]
\begin{minipage}[t]{0.52\textwidth}
   \centering
\includegraphics[width=0.8\linewidth]{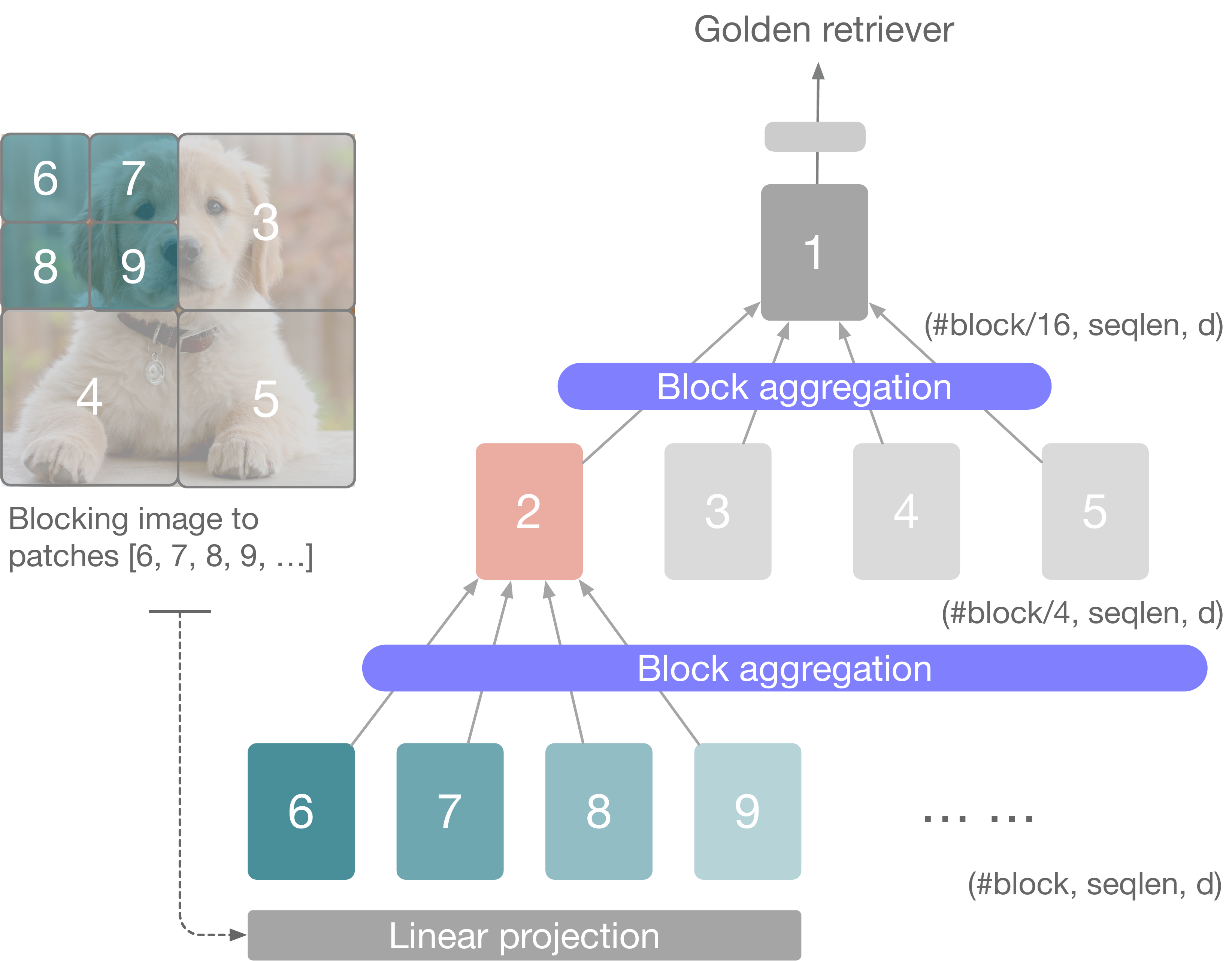} 
 
\end{minipage}
\begin{minipage}[t]{0.47\textwidth}
\vspace{-6.cm}
\makeatletter
\renewcommand{\ALG@name}{}
\renewcommand{\thealgorithm}{Pseudo code}
\makeatother
\begin{algorithm}[H]
\small
\caption{\small \OURS}
\label{alg:code}
\definecolor{codeblue}{rgb}{0.25,0.5,0.5}
\definecolor{codekw}{rgb}{0.85, 0.18, 0.50}
\lstset{
  backgroundcolor=\color{white},
  basicstyle=\fontsize{7.5pt}{7.5pt}\ttfamily\selectfont,
  columns=fullflexible,
  numbers=none,
  breaklines=true,
  captionpos=b,
  commentstyle=\fontsize{7.5pt}{7.5pt}\color{codeblue},
  keywordstyle=\fontsize{7.5pt}{7.5pt}\color{codekw},
}
\begin{lstlisting}[language=python]
# embed and block image to (#block,seqlen,d)
x = Block(PatchEmbed(input_image))

for i in range(num_hierarchy):
  # apply transformer layers T_i within each block
  # with positional encodings (PE)
  y = Stack([T_i(x[0] + PE_i[0]), ...])
  if i < num_hierarchy - 1: 
     # aggregate blocks and reduce #block by 4
    x = Aggregate(y, i)
    
h = GlobalAvgPool(x)       # (1,seqlen,d) to (1,1,d)
logits = Linear(h[0,0])    # (num_classes,)

def Aggregate(x, i):
  z = UnBlock(x)  # unblock seqs to (h,w,d)
  z = ConvNormMaxPool_i(x) # (h/2,w/2,d)
  return Block(z) # block to seqs
\end{lstlisting}
\end{algorithm}

\end{minipage}

\caption{
(Left) Illustration of \OURS with nested transformer hierarchy; (right) the simple pseudo code to generate the architecture. 
Each node T$\_$i processes an image block. 
The block aggregation is performed between hierarchies (num$\_$hierarchy$=3$ here) to achieve cross-block communication on the image (feature map) plane. }
\label{fig:arch}
\end{figure*} 

\begin{enumerate}[leftmargin=7.5mm]
    \setlength\itemsep{0.3em}
    \item We demonstrate integrating hierarchically nested transformers with the proposed block aggregation function can outperform previous sophisticated (local) self-attention variants, leading to a substantially-simplified architecture and improved data efficiency. This provides a novel perspective for achieving effective cross-block communication.
    \item \OURS achieves impressive ImageNet classification accuracy with a significantly simplified architectural design. 
    E.g., training a \OURS with 38M/68M parameters obtains $83.3\%/83.8\%$ ImageNet accuracy.The favorable data efficiency of \OURS is embodied by its fast convergence, such as achieving 75.9\%/82.3\% training with 30/100 epochs. Moreover, \OURS achieves matched accuracy on small datasets compared with popular convolutional architectures. E.g., training a \OURS with 6M parameters using a single GPU results in $96\%$ accuracy on CIFAR10 . 
    \item We show that when extending this idea beyond classification to image generation, \OURS can be repurposed into a strong decoder that achieves better performance than convolutional architectures meanwhile has comparable speed, demonstrated by $64\times 64$ ImageNet generation, which is an important to be able to adopt transformers for efficient generative modeling.
    \item Our proposed architectural design leads to decoupled feature learning and abstraction, which has significant interpretability benefits. To this end, we propose a novel method called GradCAT to interpret \OURS reasoning process by traversing its tree-like structure. This providing a new type of visual interpretability that explains how aggregated local transformers selectively process local visual cues from semantic image patches.
\end{enumerate}

\section{Related Work}
\label{sec:relatedwork}
Vision transformer-based models \cite{cordonnier2019relationship,dosovitskiy2020image} and self-attention mechanisms \cite{vaswani2021scaling,ramachandran2019stand} have recently attracted significant interest in the research community, with explorations of more suitable architectural designs that can learn visual representation effectively, such as injecting convolutional layers \cite{li2021localvit,srinivas2021bottleneck,yuan2021tokens} and building local or hierarchical structures \cite{zhang2021multi,wang2021pyramid}. 
Existing methods focus on designing a variety of self-attention modifications. Hierarchical ViT structures becomes popular both in vision \cite{liu2021swin,vaswani2021scaling} and NLP \cite{zhang2019hibert,santra2020hierarchical,liu2019hierarchical,pappagari2019hierarchical}. However, many methods  often add significant architectural complexity in order to optimize  accuracy.

One challenge for vision transformer-based models is data efficiency. 
Although the original ViT \cite{dosovitskiy2020image} can perform better than convolutional networks with hundreds of millions images for pre-training, such a data requirement is not always practical.
Data-efficient ViT (DeiT) \cite{touvron2020training,touvron2021going} attempts to address this problem by introducing teacher distillation from a convolutional network. 
Although promising, this increases the supervised training complexity, and existing reported performance on data efficient benchmarks \cite{hassani2021escaping,chen2021visformer} still significantly underperforms convolutional networks.
Since ViT has shown to improve vision tasks beyond image classification, with prior work studying its applicability to generative modeling \cite{parmar2018image,child2019generating,jiang2021transgan,hudson2021generative}, video understanding \cite{neimark2021video,akbari2021vatt}, segmentation and detection \cite{wang2020max,liang2020polytransform,kim2021hotr}, interpretability \cite{chefer2021generic,abnar2020quantifying}, a deeper understanding of the data efficiency and training difficulties from the architectural perspective is of significant impact.

\begin{figure*}[t]
\centering
\includegraphics[width=0.8\linewidth]{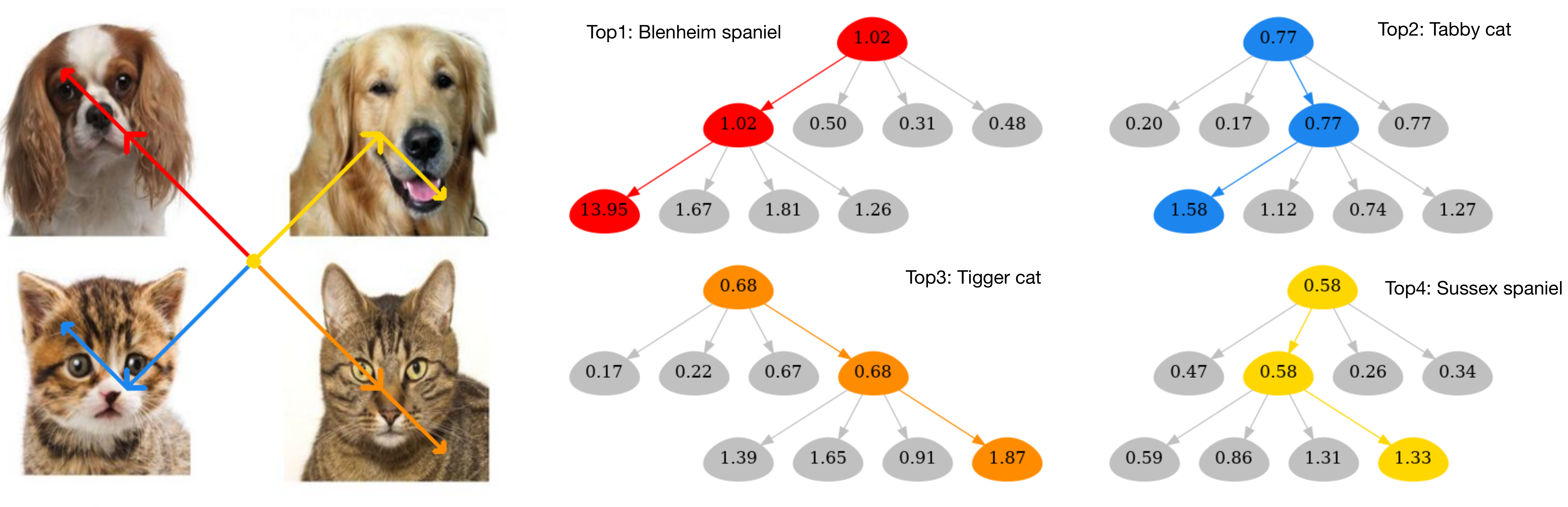} 
\caption{Example results of the proposed GradCAT. Given the left input image (containing four objects), the figure visualizes the top-4 class traversal results (4 colors) using an ImageNet-trained \OURS (with three tree hierarchies). 
Each tree node denotes the averaged activation value ($\hat{h}_l$ defined in Algorithm \ref{alg:gat}). 
The traversals can correctly find the model decision path along the tree to locate an image patch belonging to the objects of given target classes.}
\label{fig:treetrace}
\vspace{-.2cm}
\end{figure*} 

\section{Proposed Method}
\subsection{Main Architecture}
According to Fig. \ref{fig:arch}, our overall design stacks canonical transformer layers to conduct local self-attention on every image block independently, and then nests them hierarchically. 
Coupling of processed information between spatially adjacent blocks is achieved through a proposed block aggregation between every two hierarchies. 
The overall hierarchical structure can be determined by two key hyper-parameters: patch size $S\times S$ and number of block hierarchies $T_d$. 
All blocks inside each hierarchy share one set of parameters.

Given an input of image with shape $H\times W\times 3$, each image patch with size $S\times S$ is linearly projected to an embedding in $\mathbb{R}^d$. 
Then, all embeddings are partitioned to blocks and flattened to generate input $X \in \mathbb{R}^{b \times T_n \times n \times d}$, where $b$ is the batch size, $T_n$ is the total number of blocks at bottom of the \OURS hierarchy, and $n$ is the sequence length (the number of embeddings) at each block. Note that $T_n \times n = H\times W / S^2$. 

Inside each block, we stack a number of canonical transformer layers, where each is composed of a multi-head self-attention (MSA) layer followed by a feed-forward fully-connected network (FFN) with skip-connection \cite{he2016deep} and Layer normalization (LN) \cite{ba2016layer}. 
Trainable positional embedding vectors \cite{touvron2020training} are added to all sequence vectors in $\mathbb{R}^d$ to encode spatial information before feeding into the block function $T$:
\begin{equation}
\label{eq:msaours}
\begin{split}
\text{ multiple }  \times 
\begin{cases}
   y  = x + \text{MSA}_{\text{\OURS}}(x^{\prime}, x^{\prime}, x^{\prime}),  \;  x^{\prime} = \text{LN}(x)\\
   x  = y + \text{FFN}(\text{LN}(y))
 \end{cases}
\end{split}
\end{equation}
The FFN is composed of two layers: $\text{max}(0, xW_1 + b)W_2 + b$. Given input $X \in \mathbb{R}^{b \times T_n \times n \times d}$, since all blocks at one \OURS hierarchy share the same parameters, $\text{MSA}_{\text{\OURS}}$ basically MSA is applied \cite{vaswani2017attention} to all blocks in parallel:
\vspace{-.1cm}
\begin{equation} 
\label{eq:nodemsa}
\begin{split}
    \text{MSA}_{\text{\OURS}} (Q, K, V) & = \text{Stack}(\text{block}_1, ..., \text{block}_{T_n}), \\
     \text{where } \text{block}_i & = \text{MSA}(Q, K, V) W^O.
\end{split}
\end{equation}
$\text{block}_i$ has shape ${b \times n \times d}$.
Lastly, we build a nested hierarchy with block aggregation -- every four spatially connected blocks are merged into one. 
The overall design makes \OURS easy to implement, requiring minor code changes to the original ViT.

\subsection{Block Aggregation}
\label{sec:blockagg}
From a high-level view, \OURS leads to hierarchical representations, which share similarity with several pyramid designs \cite{zhang2021multi,wang2021pyramid}.
However, most of these works use global self-attention throughout the layers, interleaved with (spatial) down-sampling. 
In contrast, we show that \OURS, which leverages local attention, can lead to significantly improved data efficiency.
In local self-attention, non-local communication is important to maintain translational equivariance \cite{vaswani2021scaling}. 
To this end, Halonet \cite{vaswani2021scaling} allows the query to attend to slightly larger regions than the assigned block. 
Swin Transformer \cite{liu2021swin} achieves this by shifting the block partition windows between consecutive self-attention layers to connect adjacent blocks; applying special masked self-attention to guarantee spatial continuity. 
However, both add complexity to the self-attention layers and such sophisticated architectures are not desired from implementation perspective.

On the other hand, every block in \OURS processes information independently via standard transformer layers, and only communicate and mix global information during the block aggregation step via simple spatial operations (e.g. convolution and pooling). 
One key ingredient of block aggregation is to perform it in the image plane so that information can be exchanged between nearby blocks. 
This procedure is summarized in Fig. \ref{fig:arch}.
The output $X_l \in \mathbb{R}^{b \times {\scriptsize\#}\text{block} \times n \times d}$ at hierarchy $l$ is unblocked to the full image plane $A_{l} \in \mathbb{R}^{b \times H^{\prime} \times W^{\prime} \times d^{\prime}}$. A number of spatial operations are applied to down-sample feature maps $A_{l}^{\prime} \in \mathbb{R}^{b \times H^{\prime}/2 \times W^{\prime}/2 \times d}$. 
Finally, the feature maps are blocked back to $X_{l+1} \in \mathbb{R}^{b \times \#\text{block}/4 \times n \times d^{\prime}}$ for hierarchy $l+1$. 
The sequence length $n$ always remains the same and the total number of blocks is reduced by a factor of 4, until reduced to 1 at the top (i.e. $\#\text{block}/4^{(T_d-1)} = 1$). 
Therefore, this process naturally creates hierarchically nested structure where the ``receptive field'' expands gradually.
$d^{\prime} \ge {d}$ depends on the specific model configuration.

Our block aggregation is specially instantiated as a $3\times 3$ convolution followed by LN and a $3\times 3$ max pooling. 
Figure A2 in Appendix explains the core design and the importance of applying it on the image plane (i.e. full image feature maps) versus the block plane (i.e. partial feature maps corresponding to $2\times 2$ blocks that will be merged). 
The small information exchange through the small convolution and max. pooling kernels across block boundaries are particularly important. 
We conduct comprehensive ablation studies to demonstrate the importance of each of the design components.

Note that the resulting design shares some similarities with recent works that combine transformer and convolutional networks \cite{wu2021cvt,yuan2021tokens,bello2021lambdanetworks} as specialized hybrid structures. 
However, unlike these, our proposed method aims to solve cross-block communications in local self-attention, and the resulting architecture is simple as a stacking of basic transformer layers. 

\section{Transposed \OURS for Image Generation}
The data efficiency and straightforward implementation of \OURS makes it desirable for more complex learning tasks. 
With transpose the key ideas from \OURS to propose a decoder for generative modeling, and show that it has better performance than convolutional decoders with comparable speed. 
Remarkably, it is nearly a magnitude faster than the transformer-based decoder TransGAN \cite{jiang2021transgan}. 

Creating such a generator is straightforward by transposing \OURS (see Table A6 of Appendix for architecture details).  
The input of the model becomes a noise vector and the output is a full-sized image. 
To support the gradually increased number of blocks, the only modification to \OURS is replacing the block aggregation with appropriate block de-aggregation, i.e. up-sampling feature maps (we use pixel shuffle \cite{shi2016real}).
The feature dimensions in all hierarchies are $(b, nd) \rightarrow (b, 1, n, d) \rightarrow (b, 4, n, d^{\prime}), ..., \rightarrow (b, \#\text{blocks}, n, 3)$. 
The number of blocks increases by a factor of 4. 
Lastly, we can unblock the output sequence tensor to an image with shape $H \times W \times 3$. 
The remaining adversarial training techniques are based on  \cite{goodfellow2014generative,zhang2019self} as explained in experiments.
Analogous to our results for image classification, we show the importance of careful block de-aggregation design, in making the model significantly faster while achieving better generation quality.

\section{GradCAT: Interpretability via Tree Traversal}

Different from previous work, the nested hierarchy with the independent block process in \OURS resembles a decision tree in which each block is encouraged to learn non-overlapping features and be selected by the block aggregation. 
This unique behavior motivates us to explore a new method to explain the model reasoning, which is an important topic with significant real world impact in convnets \cite{selvaraju2017grad,sundararajan2017axiomatic}.

\setcounter{algorithm}{0} \begin{algorithm}[h]
\small
   \caption{\small GradGAT }
   \label{alg:gat}
\begin{algorithmic}
   \STATE {\bfseries Define:}  $A_l$ denotes the feature maps at hierarchy $l$. $Y_c$ is the logit of predicted class $c$. 
   $[\cdot]_{2\times2}$ indexes one of $2\times2$ partitions of input maps.
   \STATE {\bfseries Input:} $\{A_l | l=2,...,T_d\}, \alpha_{T_d} = A_{T_d}$, $P = []$
   \STATE {\bfseries Output:} The traversal path $P$ from top to bottom
    \FOR{$l = [T_d,...,2]$} 
        \STATE $h_l = \alpha_l \cdot (- \frac{\partial Y_c}{\alpha_l})$ \hspace*{0pt}\hfill \textit{\# obtain target activation maps} \\
        \STATE $\hat{h}_l = \text{AvgPool}_{2\times2}(h_l)  \in \mathbb{R}^{2\times2} $ \hspace*{0pt}\hfill  \\
        \STATE $n^*_l = \arg \max \hat{h}_l$, \; $P = P + [n^*_l]$ \hspace*{0pt}\hfill  \textit{\# pick the maximum index} \\
        \STATE $\alpha_l = A_l[n^*_l]_{2\times2}$ \hspace*{0pt}\hfill  \textit{\# obtain the partition for the index}
   \ENDFOR
\end{algorithmic}
\end{algorithm}

We present a gradient-based class-aware tree-traversal (GradCAT) method (Algorithm \ref{alg:gat}). 
The main idea is to find the most valuable traversal from a child node to the root node that contributes to the classification logits the most. Intuitively, at the top hierarchy, each of four child nodes processes one of $2\times 2$ non-overlapping partitions of feature maps $A_{T_d}$. 
We can use corresponding activation and class-specific gradient features to trace the high-value information flow recursively from the root to a leaf node. 
The negative gradient $- \frac{\partial Y_c}{A_l}$ provides the gradient ascent direction to maximize the class $c$ logit, i.e., a higher positive value means higher importance. 
Fig. \ref{fig:treetrace} illustrates a sample result.

\section{Experiments}

We first show the benefit of \OURS for data efficient learning and then demonstrate benefits for interpretability and generative modeling. Finally, we present ablation studies to analyze the major constituents of the methods.

\topic{Experimental setup.}  We follow previous work \cite{dosovitskiy2020image} to generate three architectures that have comparable capacity (in number of parameters and FLOPS), noted as tiny (\OURS-T), small (\OURS-S), and base (\OURS-B). 
Most recent ViT-based methods follow the training techniques of DeiT \cite{touvron2020training}.
We follow the settings with minor modifications that we find useful for local self-attention (see Appendix for all architecture and training details).
We do not explore the specific per-block configurations (e.g. number of heads and hidden dimensions), which we believe can be optimized through architecture search \cite{tan2019efficientnet}.

\begin{table}[t]
\small
\caption{Test accuracy on CIFAR with input size $32\times 32$. 
The compared convolutional architectures are optimized models for CIFAR. 
All transformer-based architectures are trained from random initialization with the same data augmentation. 
DeiT uses $S=2$. Swin and our \OURS uses $S=1$. 
$^{\star}$ means model tends to diverge. 
} 
\label{tab:cifar_main}
\centering
\begin{tabular}{cl|cc}
 \toprule
 Arch. base & Method  & C10 (\%) & C100 (\%) \\  \bottomrule
\multirow{2}{*}{Convolutional}    
                         & Pyramid-164-48   &   95.97   &   80.70 \\ 
                         & WRN28-10         &  95.83   &	80.75 \\ \midrule

 \multirow{7}{*}{\shortstack{Transformer 
        \\ full-attention}}   & DeiT-T     & 88.39     & 67.52 \\ 
                              & DeiT-S    & 92.44     & 69.78 \\
                              & DeiT-B   & 92.41     & 70.49 \\ \cmidrule{2-4}
                              
                              & PVT-T      & 90.51         & 69.62 \\ 
                              & PVT-S       & 92.34         & 69.79 \\ 
                              & PVT-B       & $85.05^{\star}$ & $43.78^{\star}$ \\  \cmidrule{2-4}
                              & CCT-7/$3{\times}1$    & 94.72 & 76.67 \\ \cmidrule{1-4} 
                             
  \multirow{6}{*}{\shortstack{Transformer 
 \\ local-attention}}         & Swin-T         & 94.46            & 78.07 \\
                              & Swin-S        & 94.17            & 77.01 \\
                              & Swin-B         & 94.55            & 78.45 \\ \cmidrule{2-4}
                              
                              & \OURS-T    & 96.04	         & 78.69 \\
                              & \OURS-S      & 96.97 &	81.70\\
                              & \OURS-B     & \textbf{97.20} &	\textbf{82.56}   \\\bottomrule
                             
\end{tabular}
\vspace{-.2cm}
\end{table}

\begin{table}[t]
\small
\caption{Comparison on the ImageNet dataset. 
All models are trained from random initialization. ViT-B/16 uses an image size 384 and others use 224.} 
\label{tab:imagenet}
\centering
\begin{tabular}{cl|r|c}
 \toprule
 Arch.  base                   & Method                 & \#Params        & Top-1 acc. (\%) \\  \bottomrule
 \multirow{3}{*}{Convolutional}      & ResNet-50              & 25M              &  76.2 \\
                               & RegNetY-4G                     & 21M     & 80.0        \\                                  
                              & RegNetY-16G                     & 84M         &  82.9         \\
\midrule
                         
 \multirow{3}{*}{\shortstack{Transformer 
        \\ full-attention}}   & ViT-B/16   &   86M &  77.9              \\ 
                              & DeiT-S         &  22M  & 79.8       \\
                              & DeiT-B      &  86M & 81.8      \\  \midrule
                              
  \multirow{6}{*}{\shortstack{Transformer 
 \\ local-attention}}         & Swin-T         &  29M  & 81.3                 \\
                              & Swin-S         &  50M    & 83.0                 \\
                              & Swin-B          &  88M      & 83.3                \\ \cmidrule{2-4}
                     
                              & \OURS-T      &  17M   &  81.5           \\
                              & \OURS-S      &  38M     &   83.3         \\
                              & \OURS-B     &  68M     &  \textbf{83.8}     \\     \bottomrule
                             
\end{tabular}
\vspace{-.3cm}
\end{table}

\begin{table}[t]
   \caption{Comparison on ImageNet benchmark with ImageNet-22K pre-training.} \vspace{-.1cm} \label{tab:imagenet22k} 
    \centering
    \small
    \begin{tabular}{l|c|c|c}
        \toprule
                                 &  ViT-B/16  & Swin-B  & Nest-B \\ \midrule
        ImageNet Acc. (\%)   &   84.0     &   86.0   & \textbf{86.2} \\\bottomrule
    \end{tabular} \vspace{-.3cm}
\end{table}

\subsection{Comparisons to Previous Work}
\topic{CIFAR.}
We compare \OURS to recent methods on CIFAR datasets \cite{krizhevsky2009learning} in Table \ref{tab:cifar_main},  to investigate the data efficiency. 
It is known that transformer-based methods usually perform poorly on such tasks as they typically require large datasets to be trained on.
The models that perform well on large-scale ImageNet do not necessary work perform on small-scale CIFAR, as the full self-attention based models require larger training datasets. 
DeiT \cite{touvron2020training} performs poorly and does not improve given bigger model size. 
PVT \cite{wang2021pyramid} has also a full self-attention based design, though with a pyramid structure. 
PVT-T seems to perform better than DeiT-T when model size is small, however, the performance largely drops and becomes unstable when scaling up, further suggesting that full self-attention at bottom layers is not desirable for data efficiency. 
Other transformer-based methods improve slowly with increasing model size, suggesting that bigger models are more challenging to train with less data. 
We attribute this to to their complex design (i.e. shifted windows with masked MSA) requiring larger training datasets, while   \OURS benefiting from a judiciously-designed block aggregation.
We also include comparisons with convolutional architectures that are specifically optimized for small CIFAR images and show that \OURS can give better accuracy without any small dataset specific architecture optimizations (while still being larger and slower, as they do not incorporate convolutional inductive biases).
The learning capacity and performance of \OURS get better with increased model size. 
Most variants of \OURS in Fig. A1 of Appendix outperform compared methods with far better throughput. E.g., $\text{\OURS}_3$-T ($S=2$) leads to $94.5\%$ CIFAR10 accuracy with 5384 images/s throughout, $10\times$ faster than the best compared result $94.6\%$ accuracy. More details can be found in Appendix.

\topic{ImageNet.}
We test \OURS on standard ImageNet 2012 benchmarks \cite{deng2009imagenet} with commonly used 300 epoch training on TPUs in Table \ref{tab:imagenet}. 
The input size is $224\times 224$ and no extra pre-training data is used. 
DeiT does not use teacher distillation, so it can be viewed as ViT \cite{dosovitskiy2020image} with better data augmentation and regularization.
\OURS matches the performance of prior work with a significantly more straightforward design (e.g. \OURS-S matches the accuracy of Swin-B, $83.3\%$).
The results of \OURS suggest that correctly aggregating the local transformer can improve the performance of local self-attention.

\topic{ImageNet-22K.} We scale up \OURS  to ImageNet-22K following the exact training schedules in \cite{liu2021swin,dosovitskiy2020image}. The pre-training is 90 epoch on $224{\times}224$ ImageNet21K images and finetuning is 30 epoch on $384{\times}384$ ImageNet images. 
Table \ref{tab:imagenet22k} compares the results. \OURS again achieves competitive results, with a significantly more straightforward design.

\begin{figure*}[t]
\centering
\begin{minipage}[t]{0.705\textwidth}
    \includegraphics[width=0.98\linewidth]{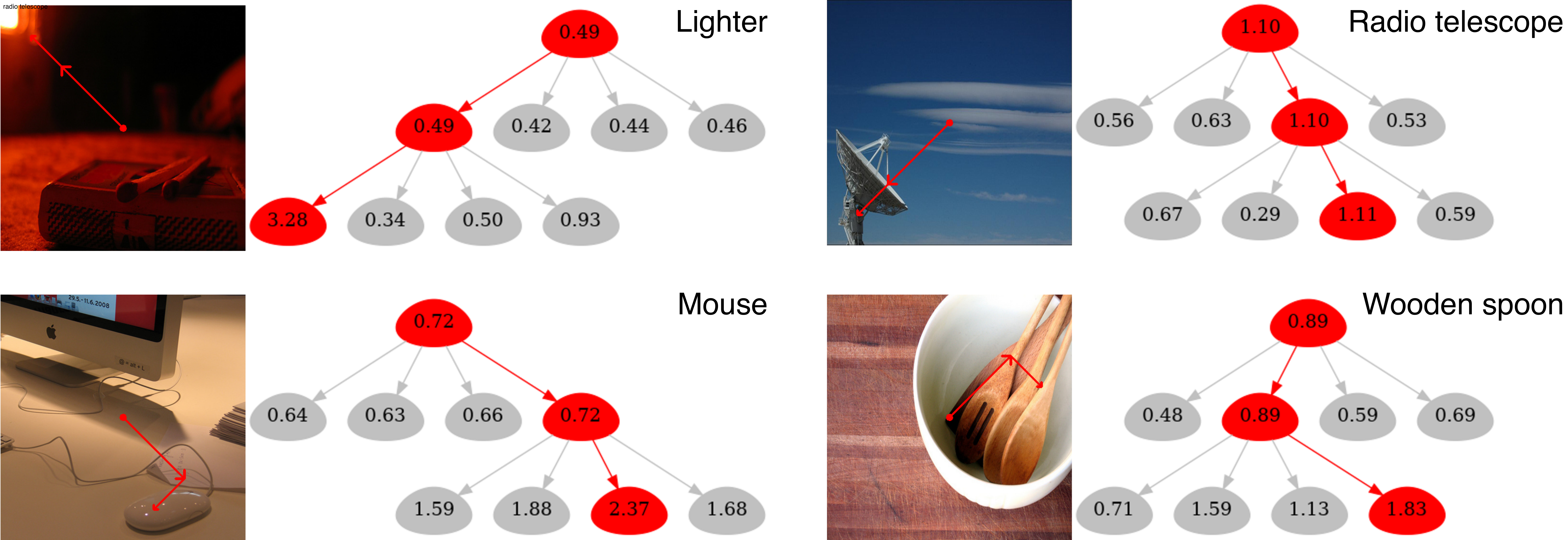} 
\end{minipage}
\begin{minipage}[t]{0.28\textwidth}
    \scriptsize
    \vspace{-3.2cm}
    \begin{tabular}{l|c}
\toprule
        Method                               & \shortstack{Top-1 loc. \\ err ($\%$)} \\ \midrule
        DeiT RollOut$^{\S}$                         & 67.6 \\
        ADL \cite{choe2019attention}         & 55.1 \\
        ACoL \cite{zhang2018adversarial}     & 54.2 \\ 
        R50 GradCAM++$^{\S}$                        & 53.5 \\ \midrule
        
        \OURS CAM                            & \textbf{51.8} \\ \bottomrule
    \end{tabular}
\end{minipage}

\caption{Left: Output visualization of the proposed GradGAT. 
Tree nodes annotate the averaged responses to the predicted class. 
We use a \OURS-S with three tree hierarchies. 
Right: CAM-based weakly supervised localization comparison on the ImageNet validation set. $^{\S}$ indicates results obtained by us. 
} \label{fig:gat}
\vspace{-.2cm}
\end{figure*}

\begin{figure*}[h] 
\centering
\includegraphics[width=0.85\linewidth]{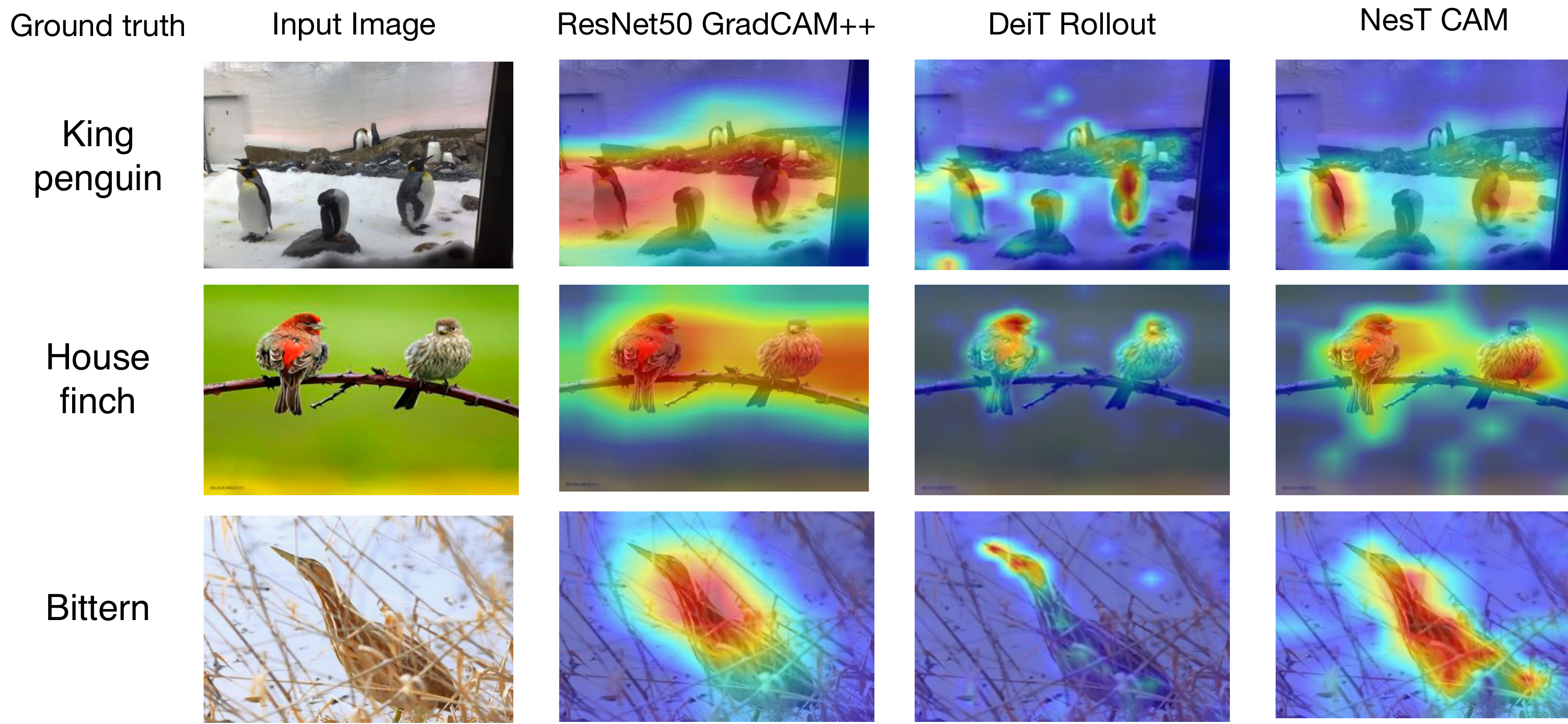} 
\caption{Visualization of CAM-based attention results. 
All models are trained on ImageNet. 
CAM (vanilla) with \OURS achieves accurate attention patterns on object regions, yielding finer attention to objects than DeiT Rollout \cite{abnar2020quantifying} and less noise than ResNet50 GradCAM++ \cite{chattopadhay2018grad}. } \vspace{-.2cm}
\label{fig:cam}
\end{figure*}

\subsection{Visual Interpretability}

\topic{GradGAT results.}
Fig. \ref{fig:gat} (left) shows the explanations obtained with the proposed GradGAT.
For GradGAT, each tree node corresponds to a value that reflects the mean activation strength. Visualizing the tree traversal through image blocks, we can get insights about the decision making process of \OURS. 
The traversal passes through the path with the highest values. As can be seen, the decision path can correctly locate the object corresponding to the model prediction.
The Lighter example is particularly interesting because the ground truth class -- lighter/matchstick -- actually defines the bottom-right matchstick object, while the most salient visual features (with the highest node values) are actually from the upper-left red light, which conceptually shares visual cues with a lighter. Thus, although the visual cue is a mistake, the output prediction is correct. This example reveals the potential of using GradGAT to conduct model diagnosis at different tree hierarchies.
Fig. A5 of Appendix shows more examples.
 
\topic{Class attention map (CAM) results.} 
In contrast to ViT \cite{dosovitskiy2020image} which uses class tokens, \OURS uses global average pooling before softmax.
This enables conveniently applying CAM-like \cite{zhou2016learning} methods to interpret how well learned representations measure object features, as the activation coefficients can be directly without approximate algorithms.
Fig. \ref{fig:gat}(right) shows quantitative evaluation of weakly-supervised object localization, which is a common evaluation metric for CAM-based methods \cite{zhou2016learning}, including GradCAM++ \cite{chattopadhay2018grad} with ResNet50 \cite{he2016deep}, DeiT with Rollout attention \cite{abnar2020quantifying}, and our \OURS CAM \cite{zhou2016learning}. 
We follow \cite{gildenblat2021explorerollout} in using an improved version of Rollout.
\OURS with standard CAM, outperforms others that are specifically designed for this task. 
Fig. \ref{fig:cam} shows a qualitative comparison (details in the Appendix), exemplifying that \OURS can generate clearer attention maps which converge better on objects.

Overall, decoupling local self-attention (transformer blocks) and global information selection (block aggregation), which is unique to our work, shows significant potential for making models easier to interpret.

\begin{figure*}[t]
\centering
\begin{minipage}[t]{0.55\textwidth}
    \centering
    \includegraphics[width=0.99\linewidth]{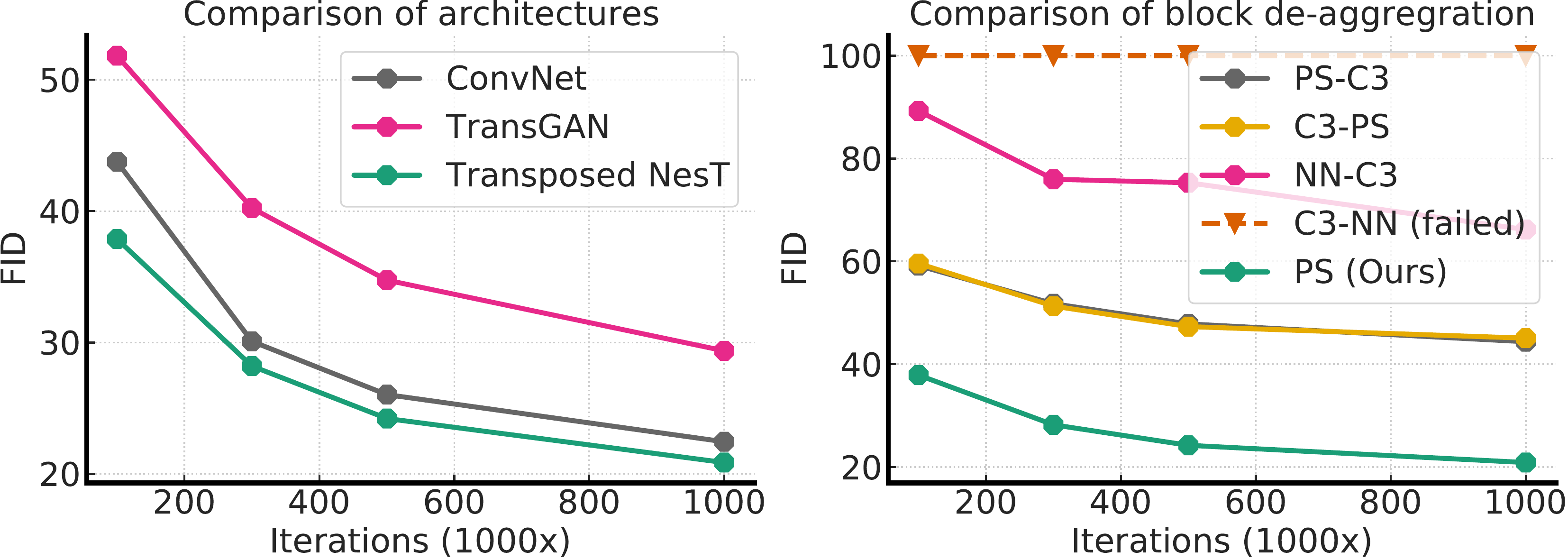} 
    
\end{minipage}
~~~
\begin{minipage}[t]{0.37\textwidth}
    \vspace{-3.4cm}
    \scriptsize
\begin{tabular}{l|cc}
\toprule
    Method      & \shortstack[r]{\#Params \\ (millions)} & \shortstack[r]{Throughput \\ (images/s)} \\ \midrule
    \shortstack[l]{Convnet \\ \cite{zhang2019self}}     &  77.8M   & 709.1 \\ \midrule
    \shortstack[l]{TransGAN \\ \cite{jiang2021transgan}}   &  82.6M   & 67.7 \\ \midrule
    \shortstack[l]{Transposed \OURS}      &  74.4M   & 523.7 \\ \bottomrule
    \end{tabular}
\end{minipage}
\caption{Left: FID comparison for $64\times 64$ ImageNet generation at different training iterations. Middle: FID comparison of different popular un-sampling methods for block de-aggregation, including combinations of pixel shuffling (PS), Conv3x3 (C3), and nearest neighbor (NN). Right: The number of parameters and throughput of compared generators.}
\label{fig:image_generation}
\vspace{-.2cm}
\end{figure*}

\subsection{Generative Modeling with Transposed \OURS}
We evaluate the generative ability of Transposed \OURS on ImageNet~\cite{russakovsky2015imagenet} where all images are resized to $64 \times 64$ resolution. 
We focus on the unconditional image generation setting to test the effectiveness of different decoders. 
We compare Transposed \OURS to TransGAN~\cite{jiang2021transgan}, that uses a full Transformer as the generator, as well as a convolutional baseline following the widely-used architecture from~\cite{zhang2019self} (its computationally expensive self-attention module is removed). 
Fig.~\ref{fig:image_generation} shows the results. 
Transposed \OURS obtains significantly faster convergence and achieves the best FID and Inception score (see Fig.~A6 of Appendix for results).
Most importantly, it achieves $8\times$ throughput over TransGAN, showing its potential for significantly improving the efficiency of transformer-based generative modeling. 
More details are explained in the Appendix.

It is noticeable from Fig. \ref{fig:image_generation} (middle) that appropriate un-sampling (or block de-aggregation) impacts the generation quality. 
Pixel shuffle \cite{shi2016real} works the best and the margin is considered surprisingly large compared to other alternatives widely-used in convnets. 
This aligns with our main findings in classification, suggesting that judiciously injecting spatial operations is important for nested local transformers to perform well.

\subsection{Ablation Studies}
We summarize key ablations below (more in Appendix).

\begin{figure}[t]
\centering
\begin{minipage}[t]{0.4\textwidth}
    \centering
    \includegraphics[width=0.8\linewidth]{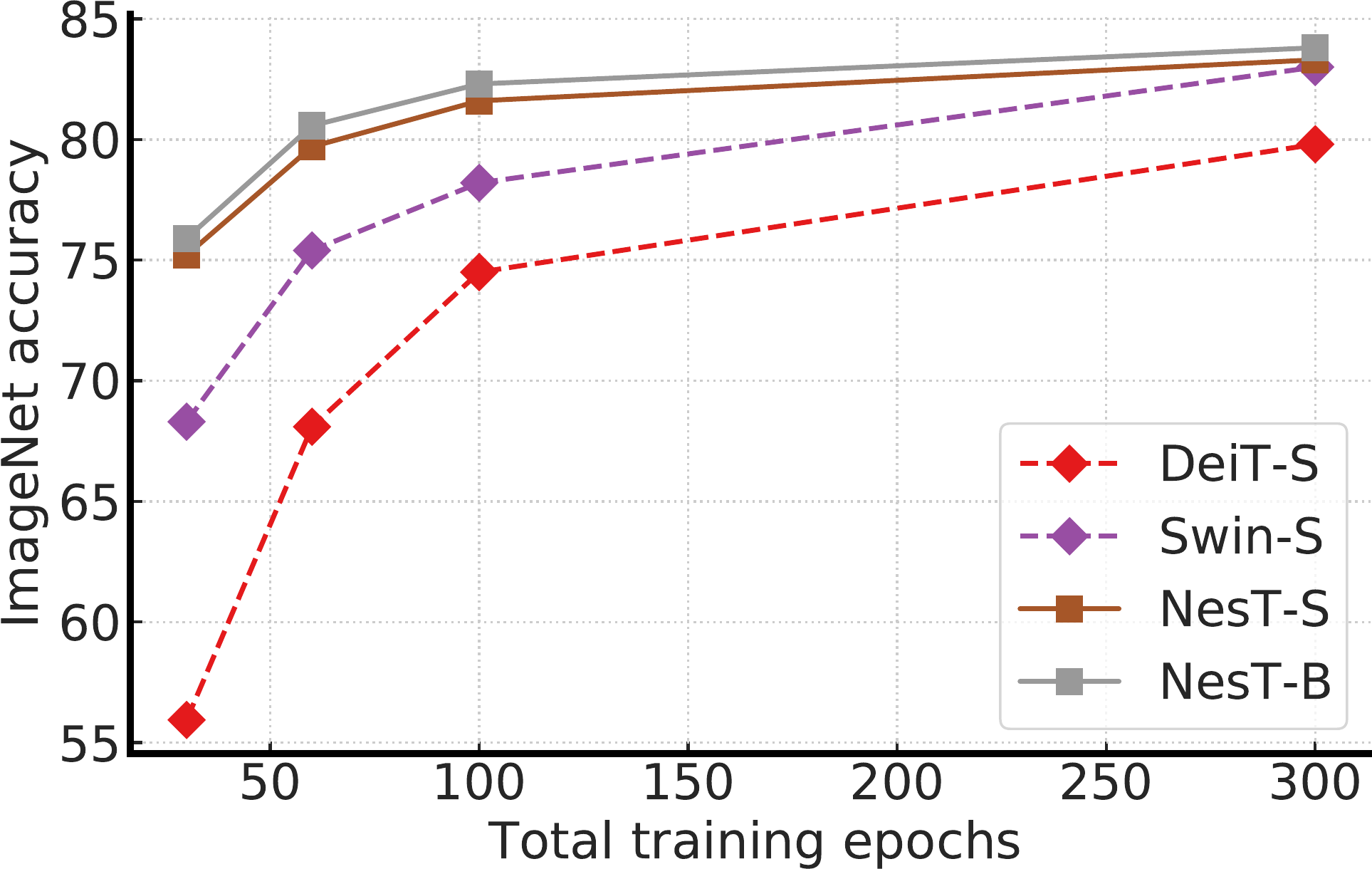} 
\end{minipage}

\vspace{.2cm}
\begin{minipage}[t]{0.49\textwidth}
    \small
\centering
\begin{tabular}{l|r|r}
        \toprule
        \multirow{2}{*}{\Centerstack{Augmentation \\ Removed}}      &  \multicolumn{2}{c}{\Centerstack{ImageNet \\ Accuracy (\%)}} \\ \cmidrule{2-3}
                                  & DeiT-B     & \OURS-T \\ \midrule
        None                      & 81.8        & 81.5 \\
        RandomErasing                    & 4.3         & 81.4 \\
        RandAugment                & 79.6        & 81.2 \\
        CutMix\&MixUp              & 75.8        & 79.8 \\ \bottomrule
    \end{tabular}
\end{minipage}
\caption{Top: Training convergence. NesT achieves better performance than DeiT with the same total epoch of training (each point is a single run). Bottom: Data augmentation ablations.
Results of DeiT-B \cite{touvron2020training} are reported by its paper. \OURS shows less reliance on data augmentation.}
\label{fig:study_aug_convergence}
\vspace{-.3cm}
\end{figure}

\topic{Fast convergence.}
\OURS achieves fast convergence, as shown in Fig. \ref{fig:study_aug_convergence} (top) for Imagenet training with $\{30, 60, 100, 300\}$ epochs. \OURS-B merely loses 1.5\% when reducing the training epoch from 300 to 100. The results suggest that \OURS can learn effective visual features faster and more efficiently.

\topic{Less sensitivity to data augmentation.} 
\OURS uses several kinds of data augmentation following \cite{touvron2020training}.
As shown in Fig. \ref{fig:study_aug_convergence} (right) and Fig.~A4, our method shows higher stability in data augmentation ablation studies compared to DeiT. 
Data augmentation is critical for global self-attention to generalize well, but reduced dependence on domain or task dependent data augmentation helps with generalization to other tasks.

\begin{figure}[t]
    \centering
    \includegraphics[width=0.999\linewidth]{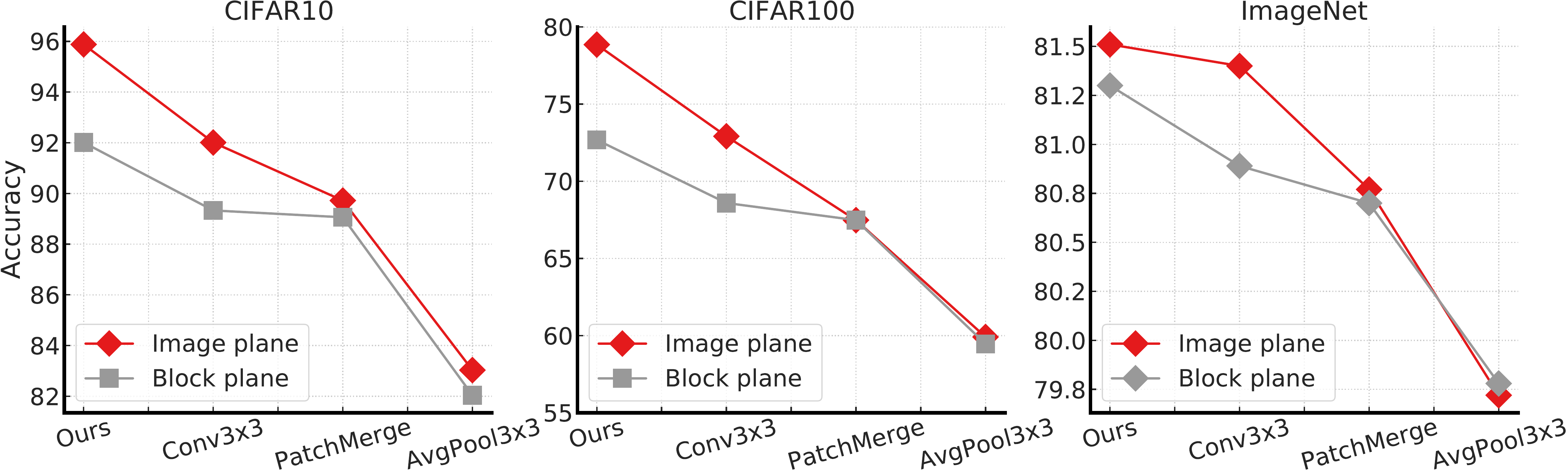} 
    \caption{Demonstration of the impact of block aggregation on CIFAR and ImageNet. \OURS-T is used. 
    Conv3x3 has stride 2. AvgPool3x3 on ImageNet is followed by Conv1x1 to change hidden dimensions of self-attention. 
    Four plausible block aggregation designs are shown in x-axis, and applied on the image plane and block plane both for comparison. Note that Ours in x-axis is Conv3x3 followed by LN and MaxPool3x3 (stride 2).
    More alternatives are validated in Fig.~A3 of Appendix.
    }
    \label{fig:node_pooling}
\vspace{-.2cm}
\end{figure}

\topic{Impact of block aggregation.} 
Here we show that the design of block aggregation is critical for performance and data efficiency.
We study this from four perspectives: 
(1) whether unblocking to the full image plane is necessary; 
(2) how to use convolution; 
(3) what kinds of pooling to use; and
(4) whether to perform query down-sampling inside self-attention \cite{vaswani2021scaling}. 
Fig. \ref{fig:node_pooling} and Fig.~A3 of Appendix compare the results of different plausible designs.

The results show that: 
(1) when performing these spatial operations, it is important to apply it on the holistic image plane versus the block plane although both can reasonably introduce spatial priors;
(2) small kernel convolution is sufficient and has to be applied ahead of pooling; 
(3) max. pooling is far better than other options, such as stride-2 sub-sampling and average pooling;
(4) sub-sampling the query sequence length (similar to performing sub-sampling on the block plane as illustrated in Fig.~A2), as used by Halonet \cite{vaswani2021scaling}, performs poorly on data efficient benchmarks. 
We also experiment PatchMerge from Swin Transformer \cite{liu2021swin} on both CIFAR and ImageNet. 
Our block aggregation closes the accuracy gap on ImageNet, suggesting that a conceptually negligible difference in aggregating nested transformers can lead to significant differences in model performance.

\section{Conclusion}
\label{sec:conclusion}
We have shown that aggregating nested transformers can match the accuracy of previous more complex methods with significantly improved data efficiency and convergence speed. 
In addition, we have shown that this idea can be extended to image generation, where it provides significant speed gains.
Finally, we have shown that the decoupled feature learning and feature information extraction in this nested hierarchy design allows for better feature interpretability through a new gradient-based class-aware tree traversal method.
In future work we plan to generalize this idea to non-image domains.

\bibliography{nips.bib}

\newpage
\appendix

\section{Appendix}
\setcounter{table}{0}
\renewcommand{\thetable}{A\arabic{table}}
\setcounter{figure}{0}
\renewcommand{\thefigure}{A\arabic{figure}}

We provide more experimental results below to complete the experimental sections of the main paper.

\subsection{\OURS Architecture and Training Details}
\label{app:details}

\begin{table*}[h]
    \centering
    \caption{Architecture details of \OURS. In each block, the structure is defined using the protocol $[d, h] \times a, b$, where
    $[d, h]$ refers to [hidden dimensions, number of heads]; $a$ refers to the number of repeated transformer layers $V$ in Equation 1 of the main paper; $b$ refers to the total number of blocks in that hierarchy.
    Tiny, Small, and Base models have different setup and they are specified below. Note that once the hierarchy is fixed. The sequence length of all blocks are consistent. Configurations of each block for CIFAR and ImageNet are different.} \label{tab:hier}
\setlength{\tabcolsep}{3.2pt}
    \renewcommand{\arraystretch}{1.3}

    \begin{tabular}{l|ccccccc}
    \toprule
    \centering
    \multirow{2}{*}{Input size}     & \multirow{2}{*}{\shortstack{Seq. \\ length}}      & \multicolumn{5}{c}{\OURS Hierarchy (Froward direction is 5 to 1)} \\ \cmidrule{3-7}
                               &              & 1  &   2  &  3             &  4           & 5                    \\ \midrule
    \multirow{4}{*}{\shortstack{$32\times 32$ \\ $S=1$}} 
                    &   & \multicolumn{5}{c}{$d = [192, 384, 768]$ and $h=[3, 6, 12]$ for model T, S, and B} \\  \cmidrule{2-7}
                    & $8\times 8$   & $[d, h] \times 4, 1$ & $[d, h] \times 4, 4$  & $[d, h] \times 4, 16$     & - & -     \\
                    & $4\times 4$   & $[d, h] \times 3, 1$ & $[d, h] \times 3, 4$  & $[d, h] \times 3, 16$     & $[d, h] \times 3, 64$ & -     \\
                    & $2\times 2$   & $[d, h] \times 2, 1$ & $[d, h] \times 2, 4$  & $[d, h] \times 2, 16$     & $[d, h] \times 2, 64$ & $[d, h] \times 2, 256$    \\ \midrule
    \multirow{2}{*}{\shortstack{$224\times 224$ \\ $S=4$}}    
                    & & \multicolumn{5}{c}{$d=[96, 96, 128]$,  $h=[3, 6, 12]$, and $k = [8, 20, 20]$ for model T, S, and B} \\  \cmidrule{2-7}
                    & $14\times 14$     & $[d, h] \times 2, 1$ & $[2d, 2h] \times 2, 4$  & $[4d, 4h] \times k, 16$     & - & -     \\

    \bottomrule
    \end{tabular}
\end{table*}

\begin{table*}[t]
\small
\caption{Test accuracy on CIFAR with input size $32\times 32$. Compared convnets are optimized models for CIFAR. All transformer based models are trained from random initialization with the same data augmentation. The number of parameters (millions), GPU memory (MB), and inference throughput (images/s) on single GPU are compared. We minimize the word size $S\times S$ for each transformer based method. DeiT uses $S=2$. Swin and our \OURS uses $S=1$. 
$^{\star}$ means models tends to diverge. 
} 
\label{app:tab:cifar_main}
\centering
\begin{tabular}{cl|rrr|cc}
 \toprule
 Arch. base & Method  & \#Params & GPU & Throughput & C10 (\%) & C100 (\%) \\  \bottomrule
\multirow{2}{*}{Convnet}    
                         & Pyramid-164-48 \cite{han2017deep}  &  1.7M  & 126M   & 3715.9      &   95.97   &   80.70 \\ 
                         & WRN28-10 \cite{zagoruyko2016wide}        &  36.5M & 202M   & 1510.8      &   95.83   &	80.75 \\ \midrule

 \multirow{7}{*}{\shortstack{Transformer 
        \\ full-attention}}   & DeiT-T \cite{touvron2020training}  &  5.3M  & 158M   & 1905.3      & 88.39     & 67.52 \\ 
                              & DeiT-S \cite{touvron2020training} &  21.3M & 356M   & 734.7       & 92.44     & 69.78 \\
                              & DeiT-B \cite{touvron2020training} &  85.1M & 873M   & 233.7       & 92.41     & 70.49 \\ \cmidrule{2-7}
                              
                              & PVT-T  \cite{wang2021pyramid}     &  12.8M          &  266M       &   1478.1 & 90.51         & 69.62 \\ 
                              & PVT-S  \cite{wang2021pyramid}     &  24.1M          &  477M       &    707.2 & 92.34         & 69.79 \\ 
                              & PVT-B  \cite{wang2021pyramid}     &  60.9M          &  990M       &    315.1  & $85.05^{\star}$ & $43.78^{\star}$ \\  \cmidrule{2-7}
                              & CCT-7/$3{\times}1$ \cite{hassani2021escaping}       &  3.7M          &  94M       &    3040.2  & 94.72 & 76.67 \\ \cmidrule{1-7} 
                             
  \multirow{6}{*}{\shortstack{Transformer 
 \\ local-attention}}         & Swin-T  \cite{liu2021swin}    &   27.5M        &   183M   & 2399.2      & 94.46            & 78.07 \\
                              & Swin-S \cite{liu2021swin}     &   48.8M        &   311M   & 1372.5      & 94.17            & 77.01 \\
                              & Swin-B \cite{liu2021swin}     &   86.7M        &   497M   & 868.3      & 94.55            & 78.45 \\ \cmidrule{2-7}
                              
                              & \OURS-T     &  6.2M          & 187M     & 1616.9      & 96.04	         & 78.69 \\
                              & \OURS-S      &  23.4M         & 411M    & 627.9  & 96.97 &	81.70\\
                              & \OURS-B      &  90.1M          & 984M     & 189.8  & \textbf{97.20} &	\textbf{82.56}   \\\bottomrule
                             
\end{tabular}
\vspace{-.2cm}
\end{table*}

\begin{table*}[t]
\small
\caption{Comparison on  the ImageNet benchmark. The number of parameters (millions), GFLOPS, and inference throughput (images/s) evaluated on a single GPU are also compared. All models are trained from random initialization without extra pre-training. } 
\label{app:tab:imagenet}
\centering
\begin{tabular}{cl|rrrr|c}
 \toprule
 Arch.  base                   & Method        & Size          & \#Params       & GFLOPS          & Throughput       & Top-1 acc. (\%) \\  \bottomrule
 \multirow{5}{*}{Convnet}      & ResNet-50 \cite{he2016deep}    & 224           & 25M        & 3.9G             & 1226.1        &  76.2 \\
                           & RegNetY-4G \cite{radosavovic2020designing}   & 224           &  21M           & 4.0G             & 1156.7    & 80.0        \\                                  
                           & RegNetY-16G \cite{radosavovic2020designing} & 224            &  84M           & 16.0G            & 334.7         &  82.9         \\
                           \midrule
                         
 \multirow{3}{*}{\shortstack{Transformer 
        \\ full-attention}}   & ViT-B/16 \cite{dosovitskiy2020image}  & 384    &   86M & 55.4G  & 85.9  & 77.9              \\ 
& DeiT-S \cite{touvron2020training}   & 224     &  22M  &  4.6G & 940.4 & 79.8       \\
                              & DeiT-B  \cite{touvron2020training}  & 224     &  86M  &   17.5G & 292.3  & 81.8      \\  \midrule
                              
  \multirow{6}{*}{\shortstack{Transformer 
 \\ local-attention}}         & Swin-T \cite{liu2021swin}   & 224        &  29M   & 4.5G    & 755.2     & 81.3                 \\
                              & Swin-S  \cite{liu2021swin} & 224         &  50M   & 8.7G    & 436.9     & 83.0                 \\
                              & Swin-B  \cite{liu2021swin}  & 224        &  88M   & 15.4G   & 278.1     & 83.3                \\ \cmidrule{2-7}
                     
                              & \OURS-T  & 224    &  17M   & 5.8G    & 633.9     &   81.5           \\
                              & \OURS-S   & 224   &  38M   & 10.4G   & 374.5     &   83.3         \\
                              & \OURS-B   & 224   &  68M   & 17.9G   & 235.8     &  \textbf{83.8}     \\     \bottomrule
                             
\end{tabular}
\vspace{-.3cm}
\end{table*}

\begin{figure*}[t]
\begin{minipage}[t]{0.75\textwidth}
    \centering
    \includegraphics[width=0.9\linewidth]{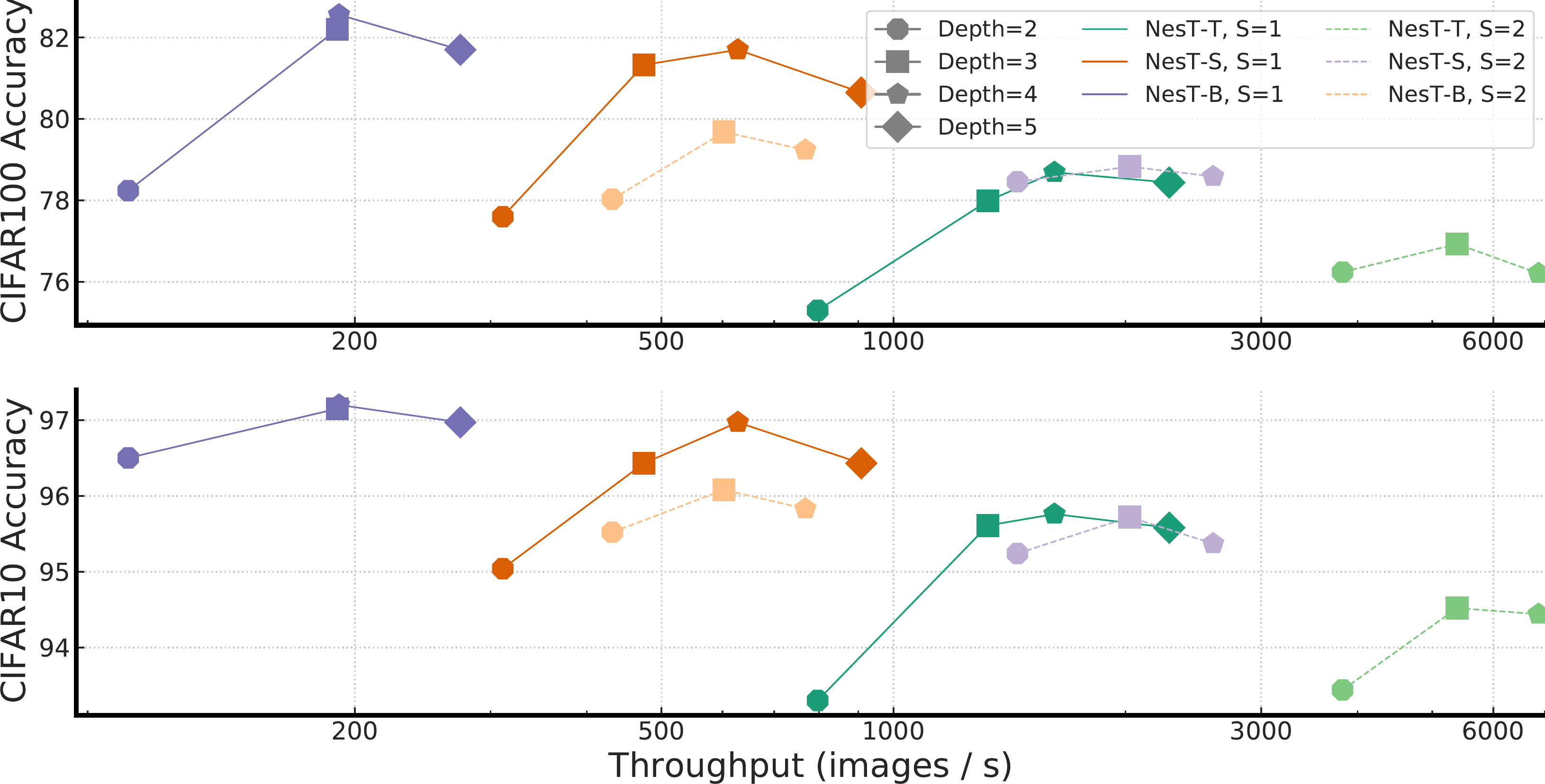} 
\end{minipage}
\begin{minipage}[t]{0.24\textwidth}
        \small
        \vspace{-4cm}
        \centering
        \setlength{\tabcolsep}{2.5pt}
        \begin{tabular}{c|c|c}
        \toprule
            \multirow{2}{*}{Depth} & \multicolumn{2}{c}{\shortstack{Node seq. \\ length $n$}}    \\ \cmidrule{2-3}
                 & $S=1$    & $S=2$   \\ \midrule
            2     & 256  & 64    \\
            3     & 64   &  16    \\
            4     & 16   &   4   \\
            5     & 4    &   -   \\ \bottomrule
    \end{tabular}
\end{minipage}
\vspace{-.1cm}
\caption{Comparison of \OURS hierarchy variants with different depth, word size $S\times S$, and model size. The right table specifies the resulting sequence length given hierarchy depth and $S$ combinations.} \label{app:fig:tree_depth}
\vspace{-.2cm}
\end{figure*}

\topic{Architecture configuration.}
The focus on this paper is how to aggregating nested transformers and its extended usage. We do not focus on the specific per-block hyper-parameters (e.g. number of heads and number of MSA layers). 
We mainly follow previous work to obtain right architectures that has similar capacity (e.g. number of parameters and throughput).

Recall that the overall hierarchy can be determined by two key hyper-parameters: patch size $S\times S$ and hierarchy depth $T_d$. 
Just like how ResNet \cite{he2016deep} adapts to small and large input sizes, \OURS also has different configuration for small input size and large input size.
We follow \cite{touvron2020training,liu2021swin} to configure the number of head, hidden dimensions for the tiny, small and base versions.
For $32\times 32$ image size, we follow \cite{touvron2020training}. Specifically, we setup the same number of repeated MSA$_{\text{\OURS}}$ per block hierarchy. In each hierarchy, the number of hidden dimensions and the number of heads are the same as well. For $224\times 224$ image size, we follow \cite{liu2021swin}. Therefore, different hierarchy has a gradually increased number of head, hidden dimensions, and number of repeated MSA$_{\text{\OURS}}$ layers. Table \ref{tab:hier} specifies details.

\subsection{\OURS Hierarchy Variants}
We study flexible variants to understand how the hierarchical structure of \OURS impacts accuracy.
When increasing the hierarchy depth by one, every four blocks are splitted to process four image partitions (see Figure 1 of the main paper).
A deeper hierarchy structure makes each block focus on a narrower range of pixel information (i.e., shorter sequence length). 

We test combinations with $S=\{1, 2\}$ and depth=$\{2,3,4,5\}$ on \OURS-\{T, S, B\}. 
Figure \ref{app:fig:tree_depth} compares different variants on two CIFAR datasets. Shallower \OURS has clear accuracy drop (although the total number of self-attention layers are the same). It is because, when depth$=1$, the model degenerates to a global self-attention method, such as ViT \cite{dosovitskiy2020image}. Depth$=5$ has marginal decrease, because the sequence length of all blocks is only $n=2\times 2$. 
Note that we denote $\text{\OURS}_{4}$-B, $S=1$ as the base \OURS with hierarchy depth 4 and word patch size $1\times1$. We use the configuration $\text{\OURS}_{4}, S=1$ as the default for the most CIFAR experiments (subscription is omitted sometimes).

\topic{Data Augmentation}. We apply the commonly used data augmentation and regularization techniques as \cite{touvron2020training}, which include a mixture of data augmentation (MixUp \cite{zhang2017mixup}, CutMix \cite{yun2019cutmix}, RandAugment \cite{cubuk2020randaugment}, RandomErasing \cite{zhong2020random}) and regularization like Stochastic Depth \cite{huang2016deep}. Repeated augmentation \cite{hoffer2020augment} in DeiT is not used.
In addition, for ImageNet models, we also add color jittering similar to~\cite{chen2020simple,chen2020big} which seems to reduce dependency on local texture cues and slightly improves generalization (${\sim}0.3\%$ on ImageNet).

\topic{Training details.} 
We use a base learning rate 2.5e-6 per device. We use the AdamW optimizer \cite{loshchilov2018fixing} and set the weight decay $0.05$. The warm-up epoch is 20. 
The initial learning rate is linearly scaled by a factor of $total\_batch\_size / 256$. 
For ImageNet training, the total batch size can be 1024 or 2048 when using distributed training on the TPU hardware. 
We use $[0.2, 0.3, 0.5]$ stochastic death rates for \OURS-T, \OURS-S, and \OURS-B models, respectively. All transformer layer weights and block aggregation weights are initialized using truncated normal distribution.

We use a 0.1 stochastic depth drop rate for all CIFAR models and the warmup is 5 epoch. The CIFAR results of compared transformer based methods, besides CCT-7/$3{\times}1$ \cite{hassani2021escaping}, in Table 1 of the main paper are trained by us. We train these models using their suggested hyper-parameters and we find it works nearly optimal on CIFAR by searching from a set of learning rate and weight decay combinations.

\begin{figure*}[t]
    \centering
    \includegraphics[width=0.9\textwidth]{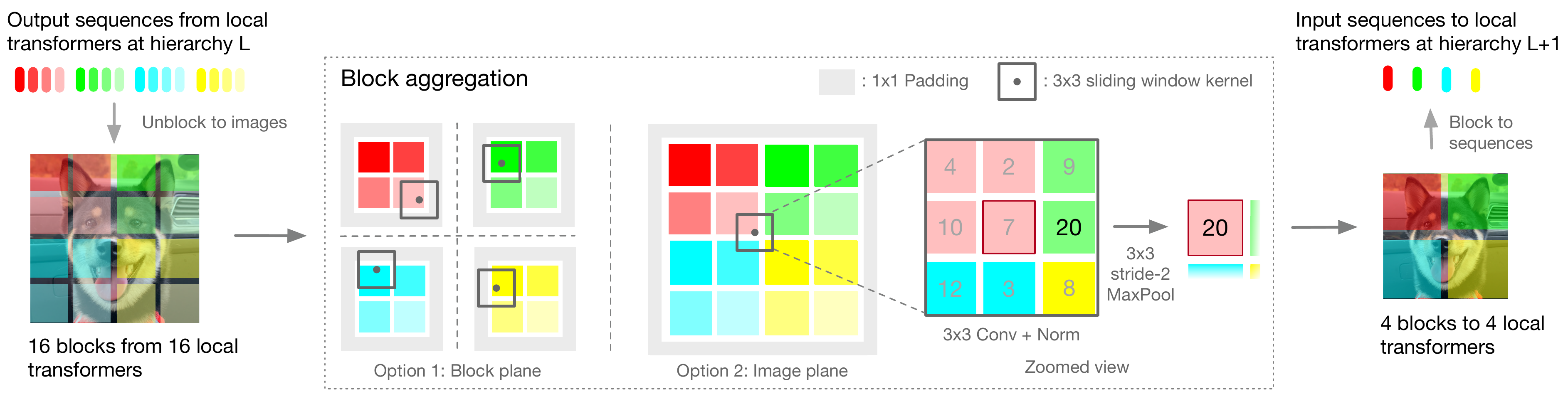}
    \caption{Illustration of block aggregation and a comparison when applying to the block plane versus on the image plane. 
    Although both perform convolution and pooling spatially, performing block aggregation on the image plane allows information communication among blocks (different color palettes) that belong to different merged blocks at the upper hierarchy.}
    \label{app:fig:block_aggregration_vis}
    \vspace{-.2cm}
\end{figure*}

\subsection{Impact of Block Aggregation}
\label{app:block}
\begin{figure*}[t]

\begin{minipage}[t]{0.19\textwidth}
        \scriptsize
        \vspace{-5cm}
        \centering
        \setlength{\tabcolsep}{2.5pt}
        \begin{tabular}{l}
        \toprule
            \shortstack[l]{Notation of  \\ spatial operations}  \\ \midrule
            C3: $3\times3$ Conv   \\
            LN: LayerNorm   \\
            MX3: $3\times3$ MaxPool   \\
            S2: $2\times2$ sub-sampling   \\
            AVG3: $3\times3$ AvgPool     \\ 
            D4: $4\times1$ 1D Conv     \\
            PM: Patch merge \\ \bottomrule
    \end{tabular}
\end{minipage}
\begin{minipage}[t]{0.8\textwidth}
    \centering
    \includegraphics[width=0.99\linewidth]{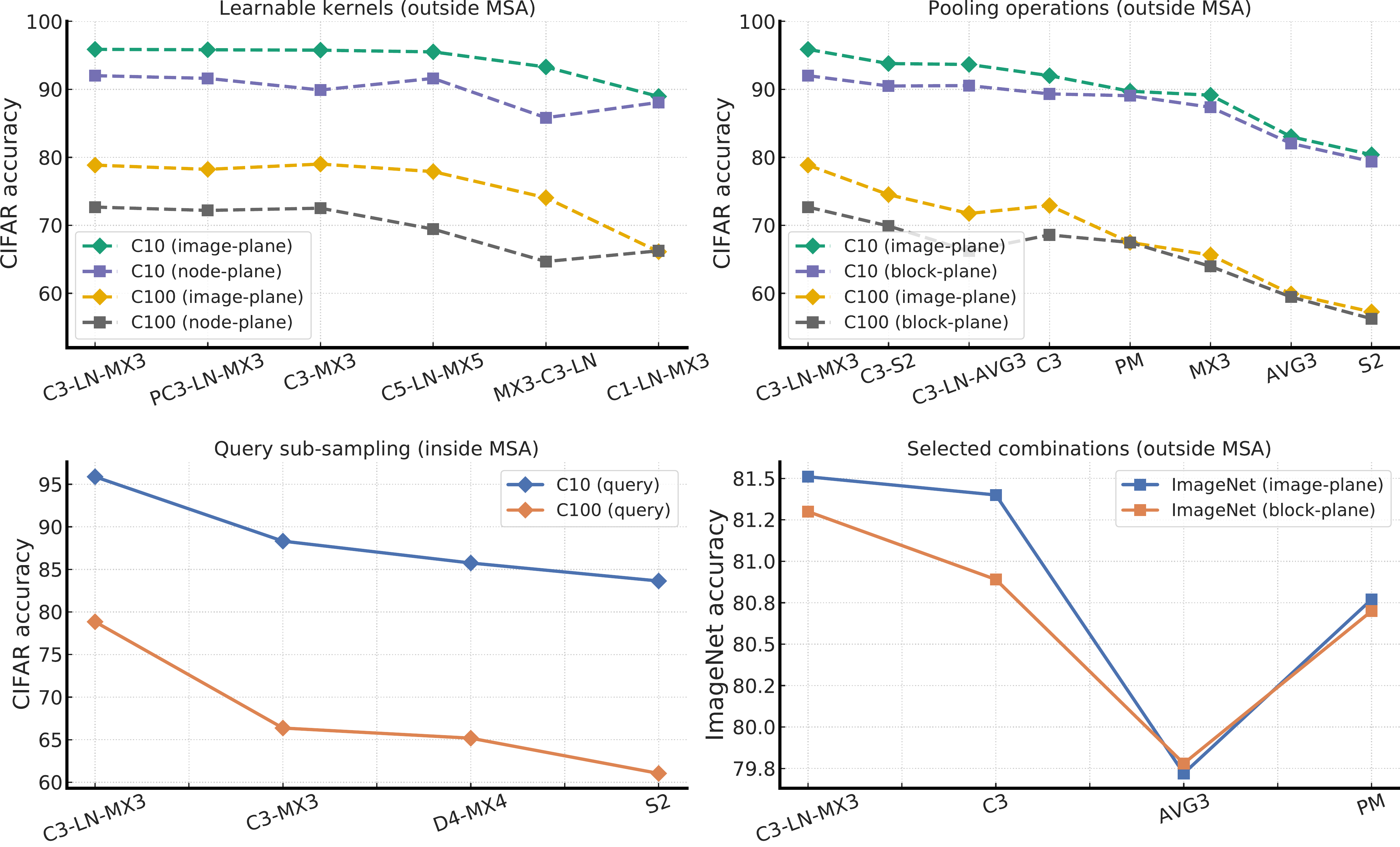} 
\end{minipage}
    \caption{Study the impact of block aggregation on CIFAR and ImageNet. \OURS-T is used.  
    We study from different perspectives as explained in the text of the main paper. 
    We verify ImageNet with \OURS-T in the bottom-right figure using a subset of representative block aggregation options found on CIFAR datasets. Patch merge \cite{liu2021swin} and 2x2 sub-sampling \cite{vaswani2021scaling} are used by previous methods. 
    Since \OURS for ImageNet has different hidden dimensions at different hierarchies, AVG3 on ImageNet is followed by a 1x1 convolution to map hidden dimensions.
    The chosen combinations are specified in x-axis. The leftmost x-axis point (C3-LN-MX3) of each figure is ours. }
    \label{app:fig:node_pooling_2}
\vspace{-.2cm}
\end{figure*}

Figure \ref{app:fig:node_pooling_2} shows detailed studies of different block aggregation functions to complete results in Figure 7 of the main paper. Although many of them has tiny difference, it is interesting that the impact to performance is non-trivial.  
In addition, we find perform query down-sampling inside self attention makes transformers more difficult to train because the skip connection also needs proper down-sampling. 

\subsection{Ablation Studies}
\label{app:studies}

\begin{figure*}[h]
    \centering
    \includegraphics[width=0.6\linewidth]{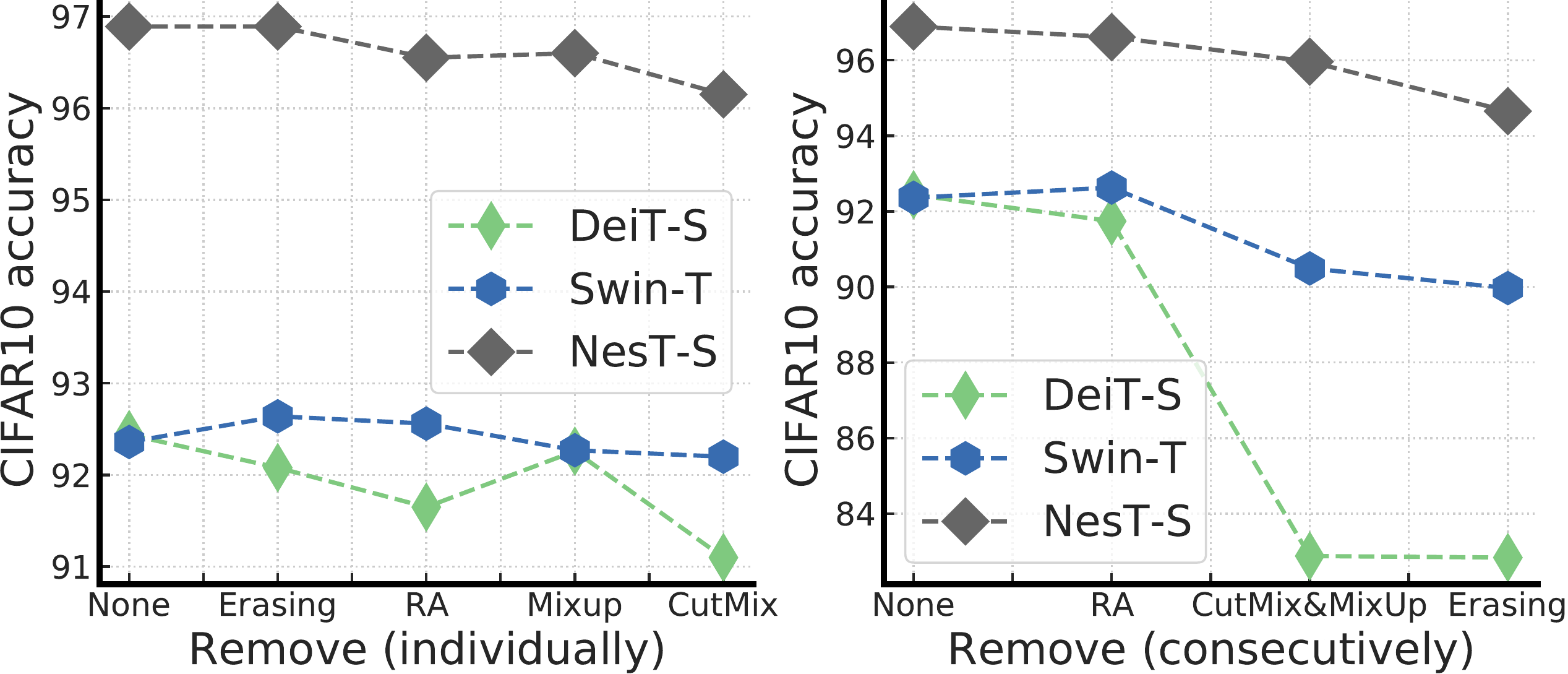} 
    \caption{Data augmentation ablation studies on CIFAR10, either removing augmentation individually (middle) or
removing (from left to right of x-axis) consecutively (right). None means all are used.}
    \label{app:fig:aug_abla_cifar}
\vspace{-.2cm}
\end{figure*}

\topic{Data augmentation sensitivity.}
\OURS uses several kinds of data augmentation types following DeiT \cite{touvron2020training}.
As shown in Figure \ref{app:fig:aug_abla_cifar}, our method shows the high stability in augmentation ablation studies compared with DeiT. We speculate the underline reason is that learning effective vision cues is much easier in local attention than in global attention, so Swin Transformer also shows comparable stability. More comparison on ImageNet will be left as future work. 

\begin{table*}[h]
    \caption{Teacher distillation studies on CIFAR datasets. Left: The top two rows of the table are teacher supervised accuracy on CIFAR100. The bottom two rows show accuracy using these trained teachers with standard or strong augmentation. Right: Teacher (PN-164-48 with standard augmentation) distillation effects on DeiT and the proposed \OURS. DeiT and \OURS are always trained with strong augmentation.} \label{app:tab:distill}
    
    \begin{minipage}[b]{0.55\textwidth}
        \vspace{.2cm}
        \setlength{\tabcolsep}{4.5pt}
        \centering
        \begin{tabular}{r|rcc}
         \toprule           
           Teacher             &   Target model    & Standard        & Strong \\ \midrule
           
               -     &      PN-164-48               &  80.7                   &  81.5     \\
               -       &      PN-164-270              & 83.4                   &  84.9          \\\midrule 
               PN-164-48       &   \OURS-B       & 84.5                   &  83.7\\
               PN-164-270         &   \OURS-B       & \textbf{84.9}        &  83.8 \\  \bottomrule
        \end{tabular}
    \end{minipage}
    \begin{minipage}[b]{0.45\textwidth}
        \setlength{\tabcolsep}{4.5pt}
        \centering
        \begin{tabular}{l|cc|cc}
         \toprule
           Distillation             &  \multicolumn{2}{c}{\xmark} & \multicolumn{2}{c}{\cmark}  \\ \midrule
           Dataset      &    C10                & C100        & C10 & C100              \\ \midrule
           DeiT-B       &   92.4   & 70.5       & 95.5       & 81.5           \\
           \OURS-B   &   \textbf{97.2}  & 82.6        & 97.1       & \textbf{84.5}           \\ \bottomrule
        \end{tabular}
    \end{minipage}
\end{table*} 

\begin{table*}[h]
    \centering
     \caption{Study the impact of number of heads in MSA on the CIFAR10 dataset with $\text{\OURS}_4$-B. When $\#$head=96, the hidden dimension used for computing Attention is only 8. However, it can still lead to similar accuracy.} \label{tab:head}
    \begin{tabular}{l|cccccccc}
    \toprule
    $\#$head  in MSA   & 1 & 2   &	3	& 6 	& 12  &	 24 &	48 &	96 \\ \midrule      
    Hidden dimension $d$    & 768 &	384	 &256 &	128	& 64 &	32&	16	&8 \\     \midrule
    Accuracy & 97.1 &	96.85&	96.92&	97.07  &	\textbf{97.21}&	97.01&	97.03&	97.08 \\ \bottomrule
    \end{tabular}
    \vspace{2pt}
   
\end{table*}

\topic{Weak teacher distillation.} 
We also explore the teacher distillation proposed by \cite{touvron2020training}, which suggests the inductive bias introduced by convnet teachers are helpful for ViT data efficiency. 
Table \ref{app:tab:distill} provides detailed distillation study on CIFAR datasets. With such a weak teacher distillation, \OURS-B is able to achieve $84.9\%$ CIFAR100 accuracy with 300 epochs and even $86.1\%$ with 1000 epoch training using 2 GPUs.

Convnet teacher distillation \cite{touvron2020training} is effective to further improve our method as well as DeiT. As explained in \cite{touvron2020training}, the inductive biases of convnets have positive implicit effects to the transformer training. 
Because data augmentation can improve teacher performance, we question the inductive biases brought by data augmentation is useful or not. Based on our experiments, it seems data augmentation negatively affect the effectiveness of teacher distillation. 
If the teacher and target model are both trained with strong augmentation, the performance decreases either for a small teacher or a big teacher.
In other words, our study suggests that training a high accuracy teacher using strong augmentation negatively impact the distillation effectiveness. Future verification on ImageNet will be left for future work.

\topic{Number of heads.} We realize different architecture design uses different number of heads for MSA. 
We attempt to understand the effectiveness of the different configurations. 
We experiment number of head from 1 to 96 given a fixed $d=768$ hidden dimension using $\text{\OURS}_4$-B.
Table \ref{tab:head} shows CIFAR10 results on \OURS. It is interesting find the number of heads affects less to the final performance. 

\begin{figure*}[h] 
\centering
\includegraphics[width=0.8\linewidth]{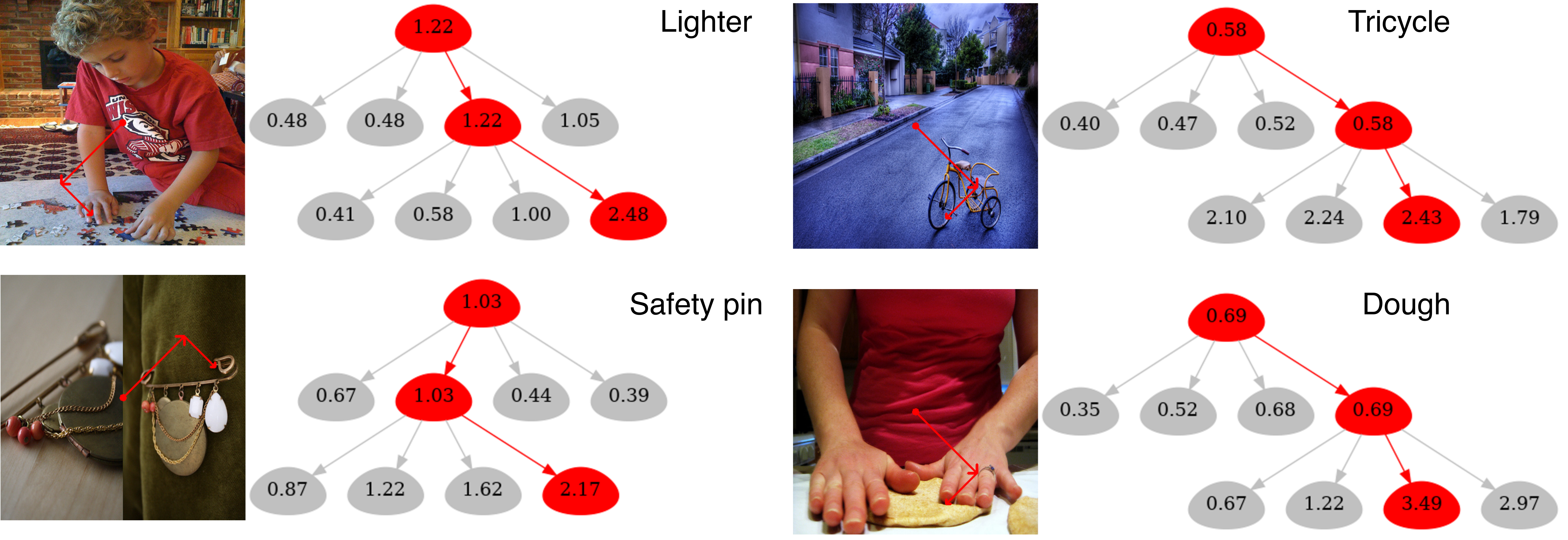} 
\caption{More output visualization of the proposed GradGAT.}
\label{app:fig:gat2}
\end{figure*} 

\subsection{Generative Modeling}
\label{app:generation}

\begin{table*}[h]
    \centering
    \caption{Architecture details of \OURS as image generator. 
    $d=1024$ and $h=4$. The input is a reshaped noise vector. At the last hierarchy, there are $64$ image blocks. Since the sequence length is $8\times 8$, it is easy to see that the output image size is $64\times 64$. At hierarchy 1, the hidden dimension is $1024/64=16$. Then a LayerNorm followed by Conv1x1 maps the hidden dimension to the output with shape $64\times64\times3$.  } \label{app:tab:hiergan}
    \renewcommand{\arraystretch}{1.3}
    \begin{tabular}{l|ccccc}
    \toprule
    \centering
    \multirow{2}{*}{Input size}     & \multirow{2}{*}{\shortstack{Seq. \\ length}}      & \multicolumn{4}{c}{\OURS Hierarchy (Froward direction is 4 to 1)} \\ \cmidrule{3-6}
                                    &                                                   & 1  &   2  &  3     &  4          \\ \midrule
    $1\times  n\times  d$ &
                     $8\times 8$  & $[d/64, h] \times 2, 64$ & $[d/16, h] \times 3, 16$  & $[d/4, h] \times 3, 4$   & $[d, h] \times 5, 1$       \\
    \bottomrule
    \end{tabular}
\end{table*}

\begin{figure*}[h]
    \centering
    \includegraphics[width=0.8\linewidth]{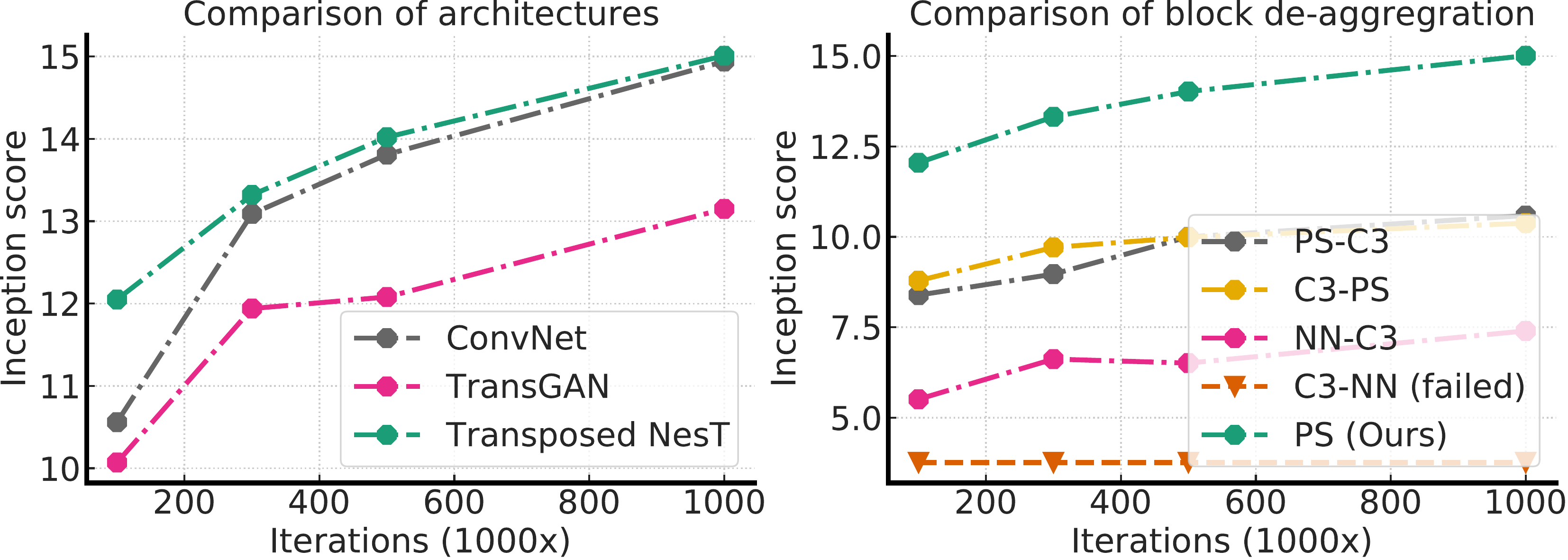} 
    \caption{Left: Inception score of $64\times 64$ ImageNet generation of different architectures. Right: Inception score with different un-pooling options.  The models used to report the results are the same models in Figure 5 of the main paper.} \label{app:fig:gan_acc_incep}
\end{figure*}

\OURS can become a decoder with minor changes.
For fair comparison in terms of model capacity, we configure \OURS following the architecture design of TransGAN for image generation.
Table \ref{app:tab:hiergan} specifies the architectural details. The block aggregation layer is swapped to a block de-aggregation layer to achieve the gradually increased image size.
Pixel shuffle (PS)~\cite{shi2016real} is leveraged to increase the image size at block de-aggregation by a factor of two while the hidden dimension is reduced to a quarter of the input. 

We adopt the same discriminator architecture as~\cite{karras2020analyzing} where R1 gradient penalty~\cite{mescheder2018training} is applied during training.  Adam~\cite{kingma2014adam} is utilized for optimization with $\beta_1 = 0$ and $\beta_2 = 0.99$. The learning rate is 0.0001 for both the generator and discriminator with mini-batches of size 256. We use Fr\'echet Inception Distance (FID)~\cite{heusel2017gans} for assessing image quality, which has been shown to be consistent with human judgments of realism~\cite{heusel2017gans,zhang2019consistency,zhang2021cross}. Lower FID values indicate closer distances between synthetic and real data distributions.
Figure \ref{app:fig:gan_acc_incep} shows the inception score of different compared methods on $64\times 64$ image generation.

\subsection{Interpretabilty}
\label{app:interp}

\topic{GradGAT results.}
GradGAT always traverses from the root node to one of the leaf node. Since each node of the bottom layer only corresponds to a small non-overlapping patch of the whole image, visualizing GradCAT is less meaningful when the targeting object is large and centered. 
We find it is true for the majority of ImageNet images although we find our results fairly promising for most ImageNet images that have small objects. 
Exploring more comprehensive studies on image datasets with non-centered objects is be left for future work. 

The proposed GradCAT is partially inspired by how GradCAM \cite{selvaraju2017grad} in convnets uses gradient information to improve visual attention. Nevertheless, the actual detailed design and serving purposes are distinct.

\topic{Class attention map results.}
Figure 4 of the main paper compares the qualitative results of CAM, including GradCAM++ \cite{chattopadhay2018grad} with ResNet50 \cite{he2016deep}, DeiT with Rollout attention \cite{abnar2020quantifying}, and our \OURS CAM \cite{zhou2016learning}. We follow \cite{gildenblat2021explorerollout} to use an improved version of Rollout, which is better than the original version. When converting CAM generated by different methods to bounding boxes, the best threshold of each method varies. We search the best threshold [0, 1] using 0.05 as the interval to find the best number for each method on the ImageNet 50k validation set. It is promising to find that \OURS CAM can outperform methods for this task and our baselines. We only use the single forward to obtain bounding boxes.

\end{document}